\documentclass[conference]{IEEEtran}
\IEEEoverridecommandlockouts
\usepackage{cite}
\usepackage{amsmath,amssymb,amsfonts}
\usepackage{graphicx}
\usepackage{adjustbox}
\usepackage[hidelinks]{hyperref}

\usepackage{textcomp}
\usepackage{xcolor}
\usepackage{url} 
\usepackage{amssymb}% http://ctan.org/pkg/amssymb
\usepackage{pifont}% http://ctan.org/pkg/pifont
\usepackage{multirow}
\usepackage{xcolor}

\usepackage{hhline}
\usepackage[linewidth=1pt]{mdframed}
\usepackage{tabularx}
\usepackage{array}
\usepackage{times}
\usepackage{longtable}
\usepackage{latexsym}
\usepackage{booktabs}
\usepackage{enumitem}
\usepackage{pifont}
\usepackage[T1]{fontenc}
\usepackage{caption}

\usepackage[utf8]{inputenc}
\definecolor{darkgreen}{RGB}{0, 150, 0} % Darker green color

\usepackage{microtype}
\usepackage[utf8]{inputenc}
\usepackage{array}
\usepackage{longtable}
\usepackage{inconsolata}
\usepackage{arydshln}

\usepackage{amssymb}
\usepackage{amsmath}
\usepackage{graphicx}
\usepackage{comment}
\usepackage{booktabs}
\usepackage{adjustbox}
\usepackage{multirow}
\usepackage{makecell}
\usepackage{algorithm}
\usepackage{algpseudocode}
\definecolor{myGreen}{RGB}{80,180,0}

\newcommand{\mypar}[1]{{{\noindent \bf #1:}}}

\usepackage{booktabs}
\def\BibTeX{{\rm B\kern-.05em{\sc i\kern-.025em b}\kern-.08em
    T\kern-.1667em\lower.7ex\hbox{E}\kern-.125emX}}
\begin{document}

\title{Full-Duplex-Bench: A Benchmark to Evaluate Full-Duplex Spoken Dialogue Models on Turn-taking Capabilities}
% \\
% % {\footnotesize \textsuperscript{*}Note: Sub-titles are not captured in Xplore and
% % should not be used}
% \thanks{}

% \author{Anonymous ASRU 2025 submission}

\author{
Guan-Ting Lin$^{1}$ \qquad Jiachen Lian$^{2*}$\thanks{$^*$Equal contribution, listed alphabetically.} \qquad Tingle Li$^{2*}$ \qquad Qirui Wang$^{3*}$ \qquad \\
Gopala Anumanchipalli$^{2}$ \qquad Alexander H. Liu$^{4}$ \qquad Hung-yi Lee$^{1}$ \\\\
\textit{$^{1}$Graduate Institute of Communication Engineering, National Taiwan University} \\
\textit{$^{2}$UC Berkeley \qquad $^{3}$University of Washington \qquad $^{4}$MIT CSAIL} \\
}

% \and
% \and
% \and
% \and

\maketitle

\begin{abstract}
Spoken dialogue modeling poses challenges beyond text-based language modeling, requiring real-time interaction, turn-taking, and backchanneling. While most Spoken Dialogue Models (SDMs) operate in half-duplex mode—processing one turn at a time—emerging full-duplex SDMs can listen and speak simultaneously, enabling more natural conversations. However, current evaluations remain limited, focusing mainly on turn-based metrics or coarse corpus-level analyses. To address this, we introduce {\em Full-Duplex-Bench}, a benchmark that systematically evaluates key interactive behaviors: pause handling, backchanneling, turn-taking, and interruption management. Our framework uses automatic metrics for consistent, reproducible assessment and provides a fair, fast evaluation setup. By releasing our benchmark and code, we aim to advance spoken dialogue modeling and foster the development of more natural and engaging SDMs.
\end{abstract}

% \footnote{Data and codebase will be released after acceptance.}


\section{Introduction}
Natural spoken dialogue is characterized by complex dynamics~\cite{schegloff2000overlapping, gravano2011turn}. Different from text-based language modeling, spoken dialogue involves problems beyond language understanding for communication, such as turn-taking~\cite{duncan1972some, gravano2011turn, raux2012optimizing}, and backchanneling~\cite{schegloff1982discourse}. These fundamental aspects facilitate mutual understanding, engagement, and social connection. Over the years, Spoken Dialogue Models (SDMs) have become a major research focus, particularly with the rise of voice assistants.

\begin{figure}[t]
  \centering
  
 \includegraphics[width=0.8\linewidth]{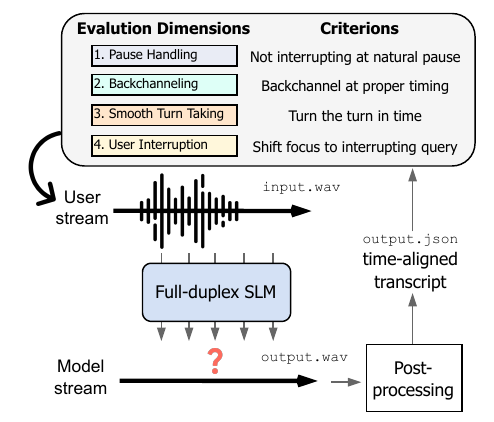}
 \caption{{\bf Overview pipeline of Full-Duplex-Bench.} We feed user audio streams to a full-duplex SDM, which produces time-synchronous output. We then perform post-processing to align both streams at the transcript level, enabling automatic evaluation along multiple dimensions.}
  \vspace{-6mm}
  \label{fig:teaser}
\end{figure}
Existing SDMs can be broadly categorized into two types: (1) {\em Half-duplex} SDMs: These models operate based on a turn-by-turn protocol, processing one audio stream at a time and switching roles in response to turn-changing signals. (2) {\em Full-duplex} SDMs: These models can simultaneously listen and speak, allowing them to capture the subtle timing of spoken conversation. They process continuous audio streams and model overlapping speech, pauses, and background noise. Furthermore, these models can capture essential spoken interaction behaviors, such as backchanneling, offering more natural and context-aware responses. 

Although half-duplex SDMs have been the dominant approach, they often sound less natural than human speakers, who seamlessly listen and speak at the same time. Recent progress in models like GPT-4o voice mode has sparked an interest in full-duplex capabilities for more human-like dialogues. Consequently, many new SDMs have emerged~\cite{syncllm, moshi, full-duplex-llm, vita, freeze-omni, minmo, wavchat}, with different architectures to enable real-time communication. As an increasing number of these systems are proposed, fair and open evaluation is critical to guide future advances in spoken dialogue research.

Evaluating SDMs is challenging because human conversation requires understanding a wide range of information and interpreting complex interactions within speech signals. Previous benchmarks have largely focused on: {\em Content-based} evaluation, measuring performance on tasks such as spoken question answering~\cite{twist, spectron, gsqa, alignslm, dual}; {\em Instruction-following} evaluation, assessing a system’s ability to follow directives~\cite{dynamic-superb, dynamic-superb-2}; and {\em Paralinguistic} evaluation, examining factors like emotion or speaking style~\cite{voxdialogue, sd-eval, air-bench, styletalk, paralingpt, emphasis}. However, these benchmarks mostly assume turn-based interactions, leaving the real-time aspects of full-duplex models under-explored.

\begin{table*}[t]
\centering
\caption{{\bf Overview of full-duplex speech language models.} ``—'' indicates properties that are not publicly available or unspecified. ``E2E'' denotes end-to-end speech modeling (without relying on text), ``\#ch'' indicates the number of speech input channels, ``Interrupt'' refers to interruption handling, ``BC'' represents backchanneling capability, and ``S2S Release'' shows whether the complete speech-to-speech pipeline is publicly released.}
\scriptsize
\setlength{\tabcolsep}{3.41mm}{
% \adjustbox{width=1\textwidth}{
\begin{tabular}{llccccc}
\toprule
\textbf{Model} & \textbf{Date} & \textbf{E2E} & \textbf{\#ch} & \textbf{Interrupt} & \textbf{BC} & \textbf{S2S Release} \\ \hline
\multicolumn{7}{l}{\textit{\textbf{Transparent Models}}} \\ 
dGSLM~\cite{dglsm} & 2022/3 & \checkmark & 2 & \checkmark & \checkmark & \checkmark \\ 
FSM~\cite{fsm} & 2024/5 & \ding{55} & 1 & - & - & \ding{55} \\ 
MiniCPM-Duplex~\cite{minicpm-duplex} & 2024/6 & \ding{55} & 1 & - & - & \ding{55} \\ 
VITA~\cite{vita} & 2024/8 & \ding{55} & 1 & \checkmark & - & \ding{55} \\ 
SyncLLM~\cite{syncllm} & 2024/9 & \checkmark & 2 & \checkmark & \checkmark & \ding{55} \\ 
Parrot~\cite{parrot} & 2024/9 & \checkmark & 2 & \checkmark & - & \ding{55} \\ 
MiniCPM-Duo~\cite{minicpm-duo} & 2024/9 & \ding{55} & 1 & - & - & \ding{55} \\ 
Moshi~\cite{moshi} & 2024/10 & \checkmark & 2 & \checkmark & \checkmark & \checkmark \\
SALMONN-omni~\cite{salmonn-omni} & 2024/11 & \checkmark & 1 & \checkmark & - & \ding{55} \\
MinMo~\cite{minmo} & 2025/1 & \checkmark & 1 & \checkmark & \checkmark & \ding{55} \\ 
OmniFlatten~\cite{omniflatten} & 2025/1 & \checkmark & 2 & \checkmark & - & \ding{55} \\ 
RTTL-DG~\cite{rttldg} & 2025/1 & \checkmark & 2 & \checkmark & \checkmark & \ding{55} \\ 
Freeze-Omni~\cite{freeze-omni} & 2024/11 & \ding{55} & 1 & \checkmark & - & \checkmark \\ \hline
\multicolumn{7}{l}{\textit{\textbf{Closed-source Commercial Models}}} \\ 
GPT-4o Voice Mode & 2024/5 & - & - & \checkmark & - & \ding{55} \\ 
Gemini Live & 2024/8 & - & - & \checkmark & - & \ding{55} \\ 
DouBao & 2025/1 & - & - & \checkmark & - & \ding{55} \\
Nova Sonic~\cite{nova-sonic} & 2025/4 & - & - & \checkmark & - & \ding{55} \\ \bottomrule
\end{tabular}}
\vspace{-5mm}
\label{tab:model}
\end{table*}

Some recent studies attempt to evaluate interaction timing in full-duplex dialogue. dGSLM~\cite{dglsm} examines voice activity patterns by comparing model outputs to ground truth gaps, interpausal units, and pauses. However, the corpus-level statistics are difficult to interpret, and the ground truth is based on a specific dialogue dataset. Recently, Talking-Turns~\cite{talking-turns} trains a specialized judge model to evaluate alignment with conversational behaviors. However, this model is trained on a specific dialogue dataset, which may limit its generalizability to other dialogue types. Moreover, the evaluation relies on user studies, making the results difficult to reproduce due to variations in participants. While these works provide valuable insights, they highlight the need for a more unified and reproducible approach to evaluating full-duplex SDMs in real-world scenarios.

% In this paper, we address the need by introducing \textbf{Full-Duplex-Bench}, the first benchmark designed to systematically evaluate key turn-taking behaviors in full-duplex spoken dialogue models (SDMs), as illustrated in Figure~\ref{fig:teaser}. Our framework targets four critical aspects of real-time interaction: {\em pause handling}, {\em backchanneling}, {\em turn-taking}, and {\em user interruption management}, using automatic metrics to enable consistent and reliable comparisons. This benchmark represents an important initial step toward evaluating the full-duplex turn-taking capabilities of SDMs. Moreover, the benchmark can be executed efficiently and automatically, enabling rapid evaluation across models. By applying our benchmark to existing systems, we gain deeper insights into how they manage interactive dynamics in real time. We will publicly release the benchmark and associated codebase to encourage further research and the development of more engaging and conversational SDMs.

In this paper, we present \textbf{Full-Duplex-Bench}, the first \emph{scenario-driven} benchmark for systematically evaluating key turn-taking behaviors in human-to-machine full-duplex spoken dialogue models (SDMs), as illustrated in Figure~\ref{fig:teaser}. The benchmark evaluates models under realistic conversational events such as backchannels and user interruptions, focusing on four critical aspects of real-time interaction: {\em pause handling}, {\em backchanneling}, {\em turn-taking}, and {\em user interruption management}. Our unified framework offers objective behavior detection, rapid diagnostic metrics, and an open, reproducible toolkit, enabling consistent and reliable cross-model comparisons. The metrics are intentionally \emph{descriptive} rather than prescriptive, allowing developers to prioritize behaviors according to specific application requirements. Designed for efficiency and full automation, the benchmark supports rapid large-scale evaluation and provides deeper insights into how existing systems manage interactive dynamics in real time. We publicly release the benchmark and codebase to facilitate further research and the development of more engaging and conversational SDMs\footnote{Data and code are released in \url{https://github.com/DanielLin94144/Full-Duplex-Bench}.}.

\section{Related Works}

\subsection{Full-duplex Spoken Dialogue Models}

Recent advances in SDMs have increasingly focused on full-duplex capabilities. We categorize existing full-duplex SDMs into two main groups: {\it Transparent Models}, which offer detailed architecture and implementation, and {\it Closed-Source Commercial Systems}, which are accessible only through demos. Table~\ref{tab:model} summarizes the key characteristics of these models.

\mypar{Transparent Models}
These models are publicly available with open-source implementations or detailed technical descriptions, enabling reproducibility and deeper analysis. They fall into two subcategories:

\noindent\textit{Cascaded Models.}
These systems follow a modular pipeline structure, typically integrating ASR, LLM, and TTS components.  
Wang \textit{et al.}~\cite{fsm} introduce control, speak, and listen tokens in an LLM framework with perceptual inputs.  
MiniCPM-Duplex~\cite{minicpm-duplex} and MiniCPM-Duo~\cite{minicpm-duo} apply time-sliced token windows for synchronous dialogue modeling.  
VITA~\cite{vita} and Freeze-Omni~\cite{freeze-omni} operate on raw speech inputs, improving latency while maintaining internal reliance on text.  
While cascaded approaches offer flexibility and stronger semantic modeling, they often suffer from cross-module latency and the loss of nuanced speech features, driving growing interest in end-to-end alternatives.

\noindent\textit{End-to-End Models.}
These models jointly model speaker and listener audio streams.  
dGSLM~\cite{dglsm} employs a Siamese network with cross-attention for two-channel dialogue.  
Moshi~\cite{moshi} uses parallel processing streams to handle real-time user speech and supports overlaps and interruptions.  
SyncLLM~\cite{syncllm} introduces time-synchronous audio chunks modeling for two-channel speech.  
OmniFlatten~\cite{omniflatten} flattens and processes speech and text tokens jointly.  
SALMONN-omni~\cite{salmonn-omni} and MinMo~\cite{minmo} inject state tokens to improve turn-taking modeling.  
Parrot~\cite{parrot} predicts next-token pairs across streams, while RTTL-DG~\cite{rttldg} uses a dialogue manager to control generation timing. By operating directly on audio data, end-to-end approaches have the potential to capture a wider range of speech features—including paralinguistic and non-verbal cues—as well as dialogue behaviors such as backchanneling.

\mypar{Closed-Source Commercial Systems}
Several commercial models have demonstrated full-duplex capabilities through public demos. These systems—such as GPT-4o Voice Mode\footnote{\url{https://openai.com/index/hello-gpt-4o/}}, Gemini Live\footnote{\url{https://blog.google/products/gemini/made-by-google-gemini-ai-updates/}}, Doubao\footnote{\url{https://team.doubao.com/en/special/realtime_voice}}, and Nova Sonic~\cite{nova-sonic}—are not open-sourced and lack publicly available architectural details.
Nonetheless, we include them in Table~\ref{tab:model}. Their emergence underscores the need for open and standardized benchmarks to evaluate interaction quality across both transparent research models and proprietary commercial systems.

\subsection{Evaluation Benchmarks}
Benchmarks offer a common framework for comparing models and understanding their performance. In the speech community, several benchmarks have been proposed to evaluate specific aspects of speech models. The SUPERB series~\cite{superb, superb-prosody, superb-sg} evaluates self-supervised speech representations across a variety of tasks. Recent advances in universal spoken language models have resulted in benchmarks that evaluate both understanding and generation. These can be grouped as follows: {\em Multi-task Instruction Following}: Dynamic-SUPERB~\cite{dynamic-superb, dynamic-superb-2}, Voicebench~\cite{voicebench}, and AIR-Bench~\cite{air-bench} test a model's ability to perform a range of tasks based on specific instructions. {\em Semantic Understanding}: For low-level language skills such as syntax and grammar, benchmarks like Zerospeech~\cite{zerospeech} and datasets such as Spoken-StoryCloze~\cite{storycloze} have been introduced. Higher-level comprehension, including spoken question answering, is measured by benchmarks like Llama-Question~\cite{spectron} and Spoken WebQuestion. {\em Paralinguistic Perception}: VoxDialogue~\cite{voxdialogue}, SD-Eval~\cite{sd-eval}, and StyleTalk~\cite{styletalk} evaluate the ability to produce natural responses by considering features such as emotion, gender, and accent. Talking-Turns~\cite{talking-turns} trains a judge model on a specific dialogue dataset to assess conversational alignment, but its generalizability is limited. Its reliance on user studies also hinders reproducibility. 
These limitations highlight the need for a unified, reproducible framework for evaluating full-duplex SDMs. To this end, we propose a benchmark with an automatic evaluation pipeline and behavior-specific metrics for assessing conversational performance.

\section{Full-Duplex-Bench Framework}

\label{subsec:pre}
For clarity, we first define several key terms that will be used in this paper:

\begin{figure*}[t]
  \centering
  \includegraphics[width=0.6\linewidth]{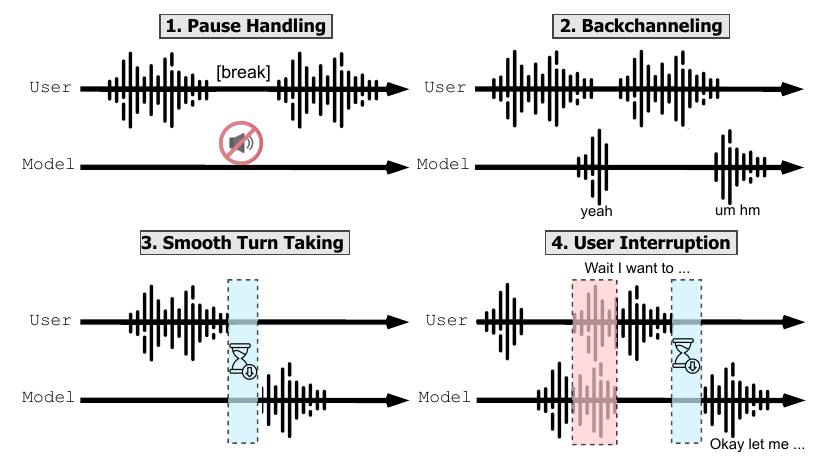}
 \caption{{\bf Illustration of the four evaluation dimensions in Full-Duplex-Bench.} (1) Pause Handling: the model stays silent during user pauses; (2) Backchanneling: the model offers short, timely acknowledgments; (3) Smooth Turn-taking: the model takes the turn in time; and (4) User Interruption: the model handles sudden user input with appropriate, well-timed responses.}
  \vspace{-5mm}
  \label{fig:detailed_dimension}
\end{figure*}

\mypar{Backchannel}
In natural conversations, listeners often provide short verbal or nonverbal signals, known as backchannels, to indicate active engagement and understanding. These backchannels include utterances such as ``mm-hmm,'' and ``uh-huh", which typically occur while another person is speaking. Effective backchanneling enhances conversational fluidity by signaling attentiveness without disrupting the speaker. Generally, backchanneling refers to short utterances produced by the listener while the speaker is talking. 

In this work, we classify a speech segment as backchanneling if it meets the following criteria: (1) it has a short duration of less than 1 second; and (2) it contains fewer than two words. This ensures that speech is delivered at a reasonable pace and that brief utterances do not interrupt the current speaker's turn. The concept of backchanneling differs across the literature~\cite{ward, reece2023candor}, so developing a comprehensive detector is beyond the scope of this paper and is left for future work.

\mypar{Takeover (TO)}
A takeover occurs when the model effectively assumes control of the conversation, dominating the turn and granting minimal opportunity for the user to speak. In this work, takeover is treated as a binary variable. If the model merely responds with silence or a backchannel, no takeover is deemed to have occurred. Conversely, any other non-silent speech that is not a backchannel indicates the model’s attempt to take over. Formally,
\begin{equation*}
    \text{TO} = \begin{cases} 0, & \text{if silence or backchannel} \\ 1, & \text{otherwise} \end{cases}
\end{equation*}

\mypar{Takeover Rate (TOR)} 
To quantify how often takeovers occur, we define the TOR as the average value of the binary TO variable across the dataset:
$\text{TOR} = \frac{1}{N} \sum_{i=1}^{N} \text{TO}_i$, where $N$ is the total number of dialogue turns, and $\text{TO}_i$ is the binary takeover variable for sample $i$.

\subsection{Overview}
As illustrated in Figure~\ref{fig:teaser}, our framework employs a unified speech input (denoted as ``\texttt{input.wav}") to simulate real-time user interactions with SDMs, enabling controlled evaluations across various dimensions. For each aspect, we design targeted test samples, collect the model's speech responses (``\texttt{output.wav}"), and use an ASR model (Nvidia \texttt{parakeet-tdt-0.6b-v2}\footnote{\url{https://huggingface.co/nvidia/parakeet-tdt-0.6b-v2}} to produce word-level, time-aligned transcriptions (``\texttt{output.json}"). Dedicated metrics are then applied to assess performance in each dimension.

\subsection{Evaluation Dimensions}  
Figure \ref{fig:detailed_dimension} illustrates the following four evaluation dimensions.

\noindent \textbf{\textit{1) Pause Handling}}

Humans naturally pause and hesitate during conversations. Pauses can occur between consecutive turns or within the same sentence. These pauses are often not intended to yield the turn but rather to maintain control of the conversation. Therefore, taking over the turn during such pauses is undesirable and may lead to user dissatisfaction. Effective pause handling ensures that the model does not interrupt when a speaker takes a natural pause.

\mypar{Research Question}
Can the model recognize when the other speaker is still holding the turn and understand that it should not take over?

\mypar{Metric}
The ideal model behavior is to avoid taking over the turn while the user is speaking. To evaluate this, we use the Takeover Rate (TOR). A lower TOR signifies better pause management, indicating that the model effectively waits for the user's turn to end. In contrast, a higher TOR suggests that the model is more likely to take over the conversation before the user has yielded the turn.

\noindent \textbf{\textit{2) Backchanneling}}

Drawing on the definition in Section~\ref{subsec:pre}, we assess whether the model, when interacting with a dominant speaker, actively listens and provides backchannels at suitable moments to facilitate dialogue engagement. A model exhibiting human-like backchanneling behavior should respond at the right times and with an appropriate frequency.

\mypar{Research Question}
Can the model determine when to offer backchannels in a human-like manner without interrupting the speaker?

\mypar{Metric}
%To measure how well models predict the timing of human backchannel cues (e.g., ``uh-huh,'' ``mm-hmm'') during conversations, we use three metrics:
To measure how well the models generate backchannel cues, we use three metrics:
\begin{itemize}[topsep=0pt, noitemsep, leftmargin=*]
    \item \textbf{TOR:} As with pause handling, the model should avoid dominating the turn, so a lower TOR is preferable. 
    \item \textbf{Backchannel Frequency (Freq):} Each backchannel event is counted and normalized by duration (events per second). When the model does not take over the turn (TOR = 0), a higher backchanneling frequency indicates that the model responds backchannel more often, but this does not necessarily imply better or more natural behavior, as it also depends on timing and context.
    \item \textbf{Jensen-Shannon Divergence (JSD):} This captures the difference between the model’s predicted timing of backchannels and actual human timing.   
 The model outputs a probability distribution \( P \), where \( P(i) \) denotes the likelihood of a backchannel occurring in time window \( i \). The ground truth distribution \( Q \) is derived from human-annotated backchannel timings (details in \ref{section:bc}), aligned to the same set of time windows. To measure the similarity between \( P \) and \( Q \), we compute the Jensen–Shannon Divergence (JSD) as:
\[
\text{JSD}(P || Q) = \frac{1}{2} \sum_{i} P(i) \log \frac{P(i)}{M(i)} + \frac{1}{2} \sum_{i} Q(i) \log \frac{Q(i)}{M(i)},
\]
where \( M(i) = \frac{1}{2}(P(i) + Q(i)) \), and \( i \) indexes the discrete time windows. JSD ranges from 0 (perfect alignment) to 1 (complete divergence), providing a symmetric and bounded measure of similarity between model predictions and human backchannel behavior. We only calculate this metric when the model does not take over the turn, where each backchannel event is counted as one-hot and normalized into a probability distribution. If the model stays silent throughout, we assume a uniform probability distribution, treating it as a random baseline without backchannel knowledge.
\end{itemize}

\noindent \textbf{\textit{3) Smooth Turn Taking}}

Effective turn-taking is crucial for maintaining a natural and engaging conversation. In human dialogue, smooth turn transitions occur when speakers respond promptly without excessive delay or overlap. A well-designed model should be capable of recognizing turn boundaries and responding with appropriate timing to ensure fluid interactions.

\mypar{Research Question}
Can the model detect the end of a speaker’s turn and respond promptly without long pauses?

\mypar{Metric}
We measure the averaged response latency, the time (in seconds) between the end of the user’s speech and the start of the model’s response. Lower latency values indicate smoother turn-taking. In cases where the model fails to respond, we record the TOR. The latency is calculated only when TO equals 1. This avoids averaging with non-takeover periods, which would introduce significant variance due to periods of silence.

\noindent \textbf{\textit{4) User Interruption}}

In human conversations, interruptions are common and can occur when a listener interjects mid-turn to clarify, disagree, or shift the discussion. A well-designed conversational model should be able to recognize and adapt to such interruptions by adjusting its response appropriately. Effective handling of interruptions ensures that the model remains responsive and coherent, even when the dialogue flow is disrupted.

\mypar{Research Question}
Can the model detect and adapt to user interruptions while maintaining a coherent and timely response?

\mypar{Metric} The evaluation consists of three measures:
\begin{itemize}[topsep=0pt, noitemsep, leftmargin=*]
\item \textbf{TOR}: TOR is measured to ensure the model takes the turn following an interruption (ideally TOR = 1). The below metrics are calculated only when the TO is 1.
\item \textbf{GPT-4o Score:} A large language model evaluates the quality of the system’s response (coherence, relevance, and adaptability) on a scale from 0 to 5. Higher scores indicate better performance. 
% Detailed instructions are provided in Appendix Table \ref{tab:gpt4_eval_prompt}.
\item \textbf{Latency After Interruption:} The averaged time taken for the model to respond after an interruption, with lower averaged latency indicating smoother interaction.
\end{itemize}

\begin{table}[t]
\centering
\caption{The number of samples for each dataset and task.}
\begin{tabular}{llc}
\toprule
\textbf{Dataset} & \textbf{Task} & \textbf{\# of Samples} \\ \midrule
Candor    & Pause Handling       & 216 \\
Candor    & Smooth Turn-Taking   & 119 \\
ICC       & Backchannel          & 55  \\
Synthetic & User Interruption    & 200 \\
Synthetic & Pause Handling       & 137 \\ \bottomrule
\end{tabular}
\vspace{-5mm}
\label{tab:dataset_stats}
\end{table}

\subsection{Data Curation}  
The overall evaluation set of Full-Duplex-Bench is shown in the Table \ref{tab:dataset_stats}. We explain how we collect the data for each evaluation dimension as follows:
% \begin{itemize}[topsep=0pt, noitemsep, leftmargin=*]

% \mypar{Candor (Pause Handling, Smooth Turn-taking)} We utilize Candor~\cite{reece2023candor}, an 850-hour dataset of open-ended, spontaneous speech conversations with two-channel recordings, as our primary data source. To construct suitable data samples for our benchmark:

% For the pause handling task, we first apply voice activity detection to identify segments with no overlapping speech, ensuring that only one speaker is active and dominant. We then select turns that contain an internal pause lasting between 0.4 and 1 second. Additionally, we filter out turns that are too short (under 5 seconds) and exclude cases where the speech before or after the pause consists of backchannel responses.

% For the smooth turn-taking task, we compute the gap between each pair of consecutive turns and retain only those where all gaps are less than 0.4 seconds. We ensure that there is no overlapping speech, each turn is longer than 4 seconds, and that the speech is not backchanneling. To allow the evaluated model sufficient time to respond, we append 5 seconds of silence at the end of the input stream.
% After automatic filtering, we manually review the audio samples to ensure quality and consistency.

\mypar{Candor (Pause Handling, Smooth Turn-taking)}
We utilize Candor~\cite{reece2023candor}, an 850-hour dataset of open-ended, spontaneous speech conversations with two-channel recordings, as our primary data source. To construct suitable data samples for our benchmark:

For the \emph{pause handling} task, we first apply voice activity detection to identify segments with no overlapping speech, ensuring that only one speaker is active. We then select turns containing an internal pause between 0.4 and 1.0 seconds. This range is informed by prior research showing that pauses of around one second are perceived as a natural upper bound in English conversation~\cite{jefferson1989preliminary}, and that pauses within this range can influence listener impressions~\cite{liu2022pause}. We exclude turns shorter than 5 seconds and remove cases where the speech immediately before or after the pause consists of backchannel responses. For the \emph{smooth turn-taking} task, we compute the gap between each pair of consecutive turns and retain only those where all gaps are less than 0.4 seconds. This threshold is based on findings that smooth conversational transitions in English typically occur within 200–250 milliseconds~\cite{heldner2010pauses}, and we conservatively set 0.4 seconds as an upper bound. We further require no overlapping speech, each turn to be longer than 4 seconds, and that the utterances are not backchanneling. To give evaluated models sufficient time to respond, we append 5 seconds of silence at the end of the input stream.

After automatic filtering, all retained segments are manually reviewed with dual-channel evidence (whether the partner is silent or entering, and whether there is hesitation or fluent continuation).

\label{section:bc}
\mypar{ICC (Backchannel)} Umair \textit{et al.}~\cite{icc} use the In Conversation Corpus (ICC) to collect Transition Relevance Places (TRPs), which are points in a speaker’s utterance that signal appropriate moments for the listener to respond. This dataset consists of 28.33 minutes of high-quality informal American English dialogues, capturing responses from 118 native speakers with no prior expertise in turn-taking research. Participants were instructed to provide brief backchannel responses (e.g., “hmm,” “yes”) whenever they found it appropriate. Their responses were recorded on individual audio channels, synchronized with the stimulus audio. On average, 59 participants responded to each stimulus turn. This allows for estimating both the likelihood of perceiving a TRP at a given moment and the distribution of these perceived response locations. By segmenting the audio into 200-ms time windows and normalizing the backchannel counts, we generate a ground truth backchannel distribution $Q$ for computing JSD for each stimulus. 

\mypar{Synthetic Data (User Interruption, Pause Handling)} To address the scarcity of user interruptions in existing public datasets, we generated synthetic dialogues using GPT-4o~\cite{gpt4}, which include both contextual turns and interruption turns. Text-to-Speech synthesis was performed using ChatTTS\footnote{\url{https://github.com/2noise/ChatTTS}}, with 10 different speaker voices randomly assigned across samples to enhance diversity. For each input stream fed to the evaluated model, the interrupting speech is played around 7 seconds after the preceding utterance. Additionally, we append 15 seconds of silence after the interruption to allow the model time to respond. 

For the synthetic \textit{pause handling} task, we leverage the \texttt{[uv\_break]} tag supported by ChatTTS to insert controlled pauses into the speech, enabling us to evaluate each model’s robustness to intra-turn pauses. In total, we collect 200 samples for user interruption and 137 samples for pause handling.

% Prompts for generating these user interruption data are provided in Appendix Table \ref{tab:generate_user_interruption}. 
% The prompt for generating pause handling data is also found in Appendix Table \ref{tab:generate_user_pause}.
% \end{itemize}

\begin{table*}[t]
    \centering
    \caption{{\bf Models comparison.} We evaluate several models across different conversational dimensions, where Latency is presented in seconds.}
    \setlength{\tabcolsep}{1.13mm}{
    \scriptsize
    \begin{tabular}{ccccccccccc}
        \toprule
        \multicolumn{1}{c}{\textbf{Dimension}} & \multicolumn{2}{c}{\textbf{Pause Handling}} & \multicolumn{3}{c}{\textbf{Backchannel}} & \multicolumn{2}{c}{\textbf{Smooth Turn Taking}} & \multicolumn{3}{c}{\textbf{User Interruption}} \\
        \cmidrule(lr){2-3} \cmidrule(lr){4-6} \cmidrule(lr){7-8} \cmidrule(lr){9-11}
        \textbf{Data} & Synthetic & Candor & \multicolumn{3}{c}{ICC} & \multicolumn{2}{c}{Candor}  & \multicolumn{3}{c}{Synthetic} \\
        \textbf{Metric} & TOR ($\downarrow$) & TOR ($\downarrow$) & TOR ($\downarrow$) & Freq ($\uparrow$) & JSD ($\downarrow$) & TOR ($\uparrow$) & Latency ($\downarrow$) & TOR ($\uparrow$) & GPT-4o($\uparrow$) & Latency ($\downarrow$) \\
        \midrule
        dGSLM & 0.934 & 0.935  & 0.691 &\textbf{0.015}  &\textbf{0.934}  &  \textbf{0.975}& 0.352  & 0.917 & 0.201 & 2.531 \\
        Moshi & 0.985 & 0.980 & 1.000 & 0.001 & 0.957 & 0.941 & \textbf{0.265}& \textbf{1.000} & 0.765 & \textbf{0.257} \\
        Freeze-Omni & \textbf{0.642} & \textbf{0.481} & \textbf{0.636} & 0.001 & 0.997 & 0.336 & 0.953 & 0.867 &\textbf{3.615} & 1.409 \\ 
        \textcolor{gray}{Gemini Live} & \textcolor{gray}{0.255} & \textcolor{gray}{0.310} & \textcolor{gray}{0.091} & \textcolor{gray}{0.012} & \textcolor{gray}{0.896} & \textcolor{gray}{0.655} & \textcolor{gray}{1.301} & \textcolor{gray}{0.891} & \textcolor{gray}{3.376} & \textcolor{gray}{1.183} \\
        \bottomrule
    \end{tabular}}
    \vspace{-5mm}
    \label{tab:performance}
\end{table*}

\section{Models Under Evaluation}  
Since many models do not release complete speech-to-speech checkpoints (see Table~\ref{tab:model}, column ``S2S Release''), currently we evaluate models with a \textbf{publicly available speech-to-speech model and inference pipeline}. We also include the recently released Gemini Live API as a representative commercial model due to its user-friendly interface.

% \footnote{By open-sourcing our data and metrics, we encourage the broader community, including those with closed-source models, to evaluate their systems using our benchmark}
\begin{itemize}[topsep=0pt, noitemsep, leftmargin=*]

\item {\bf dGSLM}~\cite{dglsm}{\bf :} A textless speech-to-speech model that generates natural conversations directly from audio. It uses (1) a HuBERT+k-means encoder~\cite{hubert}, (2) a dual-tower Transformer for two-channel dialogue modeling, and (3) a HiFi-GAN decoder. Trained on 2,000 hours of phone calls~\cite{cieri2004fisher}, it captures both linguistic and paralinguistic cues. Originally non-interactive, we adapt it for live interaction using the official implementation\footnote{\url{https://github.com/facebookresearch/fairseq/tree/main/examples/textless_nlp/dgslm}} with our modifications.
% in the Appendix.

\item {\bf Moshi}~\cite{moshi}{\bf :} A real-time speech-to-speech system combining a 7B LLM (Helium) and neural codec (Mimi) via residual vector quantization. It features an “Inner Monologue” step to improve fluency and supports overlapping speech and interruptions through a multi-stream architecture. We use the official implementation\footnote{\url{https://github.com/kyutai-labs/moshi}}.
% setup details are in the Appendix.

\item {\bf Freeze-Omni}~\cite{freeze-omni}{\bf :} A cascaded full-duplex system with a frozen LLM pipeline. VAD triggers chunk-wise encoding, and a classification head predicts dialogue states to control turn-taking. Parallel modules handle streaming, speaking, and monitoring. Evaluated locally using the official server.

\item {\bf Gemini Live}: Based on the official API documentation\footnote{\url{https://ai.google.dev/gemini-api/docs/live}}, we use the \texttt{gemini-2.0-flash-live-001} model. The \texttt{input.wav} file is first converted to 16 kHz PCM-16 format, then divided into 30 ms chunks and streamed to the Gemini Live API. Server-side voice activity detection (VAD) is used to segment the audio and trigger responses. A new session is initiated after each model reply. All generated outputs are aligned with the original input duration, preserving silence in regions where no response is produced.

\end{itemize}

\section{Results}
Table \ref{tab:performance} presents the results across four evaluation dimensions. The key findings are summarized as follows:

\mypar{Avoiding Interruptions During Speaker Pauses}
All three SDMs exhibit high Takeover Rates (TOR) when managing speaker pauses, indicating frequent interruptions during natural breaks. The end-to-end models (dGSLM and Moshi) tend to interrupt more often across both real and synthetic datasets. In contrast, Freeze-Omni, which incorporates a dedicated module for predicting speaking and listening states, demonstrates a significantly lower TOR. This suggests that an explicit turn-taking control module can better manage pauses and may offer improvements if integrated into end-to-end systems. Conversely, Gemini Live achieves the lowest TOR among all models, including open-source ones, and exhibits a higher likelihood of taking over on Candor compared to synthetic data.

\mypar{Backchanneling Dynamics}
We evaluate using TOR, backchannel frequency (Freq), and Jensen–Shannon Divergence (JSD). Similar to pause handling, Moshi frequently takes over the turn, resulting in a high TOR. Both dGSLM and Freeze-Omni achieve lower TORs. Among the three open-sourced SDMs, dGSLM produces the most backchannel responses with more natural timing (Freq = 0.015, JSD = 0.934) when it does not take over the turn. In contrast, Freeze-Omni remains largely silent, producing few backchannel responses.  The commercial Gemini Live achieves the lowest TOR and the best JSD, indicating superior ability in identifying appropriate moments for backchanneling.

\mypar{Latency of Turn-Taking}
We assessed TOR and average response latency on the Candor dataset. dGSLM and Moshi exhibit high TORs and respond quickly, with an average latency of around 0.3 seconds. Freeze-Omni, due to its cascaded architecture—which first generates text and then synthesizes speech—exhibits higher latency. Its lower TOR likely reflects missed opportunities to take over the turn, possibly due to failures in detecting turn ends. Interestingly, Gemini Live achieves a TOR of only 0.655, suggesting that even commercial models sometimes fail to take the turn in real dialogue data.

\mypar{Managing User Interruptions}
Freeze-Omni handles user interruptions effectively, achieving significantly higher contextual relevance scores while maintaining acceptable latency. This strength is attributed to its 'model-as-a-server' strategy, which leverages a pool of models to manage user barge-ins efficiently. Gemini Live shows comparable performance, with relatively better TOR and latency, though it yields a slightly lower GPT-4o score.
In contrast, the end-to-end SDMs struggle with coherence. Moshi responds promptly but yields a lower GPT-4o score (0.765), while dGSLM performs poorly, with high latency and diminished content quality (GPT-4o score: 0.201). These results highlight the challenges end-to-end systems face in preserving semantic coherence during user interruptions. 

\section{Conclusion}
In this paper, we introduced \textbf{Full-Duplex-Bench}, a benchmark designed to evaluate critical aspects of full-duplex spoken dialogue models. Our framework targets key interaction dimensions—pause handling, backchanneling, smooth turn-taking, and user interruption management—addressing limitations of existing benchmarks that primarily focus on half-duplex settings or coarse corpus-level metrics. To ensure systematic and reproducible evaluation, we propose automatic metrics tailored to real-time interaction. Experiments on full-duplex models reveal distinct model features and highlight areas for improvement. By releasing our data and behavior-specific metrics, we hope \textbf{Full-Duplex-Bench} provides a practical foundation for evaluating full-duplex spoken dialogue systems.

\section{Limitation and Future Work}
Our framework does not yet link described behaviors to human preferences; users need to determine what constitutes desirable or undesirable behavior according to their specific goals. Future work can integrate human judgment studies to provide the preference. The present analysis is limited to English, and extending the framework to other languages will be essential for assessing cross-linguistic generality.

% \section*{Acknowledgment}

% \section*{References}
% \newpage

\bibliographystyle{IEEEtran}
\footnotesize
\bibliography{refs}

\begin{thebibliography}{10}
\providecommand{\url}[1]{#1}
\csname url@samestyle\endcsname
\providecommand{\newblock}{\relax}
\providecommand{\bibinfo}[2]{#2}
\providecommand{\BIBentrySTDinterwordspacing}{\spaceskip=0pt\relax}
\providecommand{\BIBentryALTinterwordstretchfactor}{4}
\providecommand{\BIBentryALTinterwordspacing}{\spaceskip=\fontdimen2\font plus
\BIBentryALTinterwordstretchfactor\fontdimen3\font minus \fontdimen4\font\relax}
\providecommand{\BIBforeignlanguage}[2]{{%
\expandafter\ifx\csname l@#1\endcsname\relax
\typeout{** WARNING: IEEEtran.bst: No hyphenation pattern has been}%
\typeout{** loaded for the language `#1'. Using the pattern for}%
\typeout{** the default language instead.}%
\else
\language=\csname l@#1\endcsname
\fi
#2}}
\providecommand{\BIBdecl}{\relax}
\BIBdecl

\bibitem{schegloff2000overlapping}
E.~A. Schegloff, ``Overlapping talk and the organization of turn-taking for conversation,'' \emph{Language in society}, vol.~29, no.~1, pp. 1--63, 2000.

\bibitem{gravano2011turn}
A.~Gravano and J.~Hirschberg, ``Turn-taking cues in task-oriented dialogue,'' \emph{Computer Speech \& Language}, vol.~25, no.~3, pp. 601--634, 2011.

\bibitem{duncan1972some}
S.~Duncan, ``Some signals and rules for taking speaking turns in conversations.'' \emph{Journal of personality and social psychology}, vol.~23, no.~2, p. 283, 1972.

\bibitem{raux2012optimizing}
A.~Raux and M.~Eskenazi, ``Optimizing the turn-taking behavior of task-oriented spoken dialog systems,'' \emph{ACM Transactions on Speech and Language Processing (TSLP)}, vol.~9, no.~1, pp. 1--23, 2012.

\bibitem{schegloff1982discourse}
E.~A. Schegloff, \emph{Discourse as an interactional achievement: Some uses of" uh huh" and other things that come between sentences}.\hskip 1em plus 0.5em minus 0.4em\relax Analyzing discourse: Text and talk/Georgetown University Press, 1982.

\bibitem{syncllm}
B.~Veluri, B.~Peloquin, B.~Yu, H.~Gong, and S.~Gollakota, ``Beyond turn-based interfaces: Synchronous llms as full-duplex dialogue agents,'' in \emph{Proceedings of the 2024 Conference on Empirical Methods in Natural Language Processing}, 2024, pp. 21\,390--21\,402.

\bibitem{moshi}
\BIBentryALTinterwordspacing
A.~D\'efossez, L.~Mazar\'e, M.~Orsini, A.~Royer, P.~P\'erez, H.~J\'egou, E.~Grave, and N.~Zeghidour, ``Moshi: a speech-text foundation model for real-time dialogue,'' Kyutai, Tech. Rep., September 2024. [Online]. Available: \url{http://kyutai.org/Moshi.pdf}
\BIBentrySTDinterwordspacing

\bibitem{full-duplex-llm}
\BIBentryALTinterwordspacing
P.~Wang, S.~Lu, Y.~Tang, S.~Yan, W.~Xia, and Y.~Xiong, ``A full-duplex speech dialogue scheme based on large language model,'' in \emph{The Thirty-eighth Annual Conference on Neural Information Processing Systems}, 2024. [Online]. Available: \url{https://openreview.net/forum?id=YawXY6mWiK}
\BIBentrySTDinterwordspacing

\bibitem{vita}
C.~Fu, H.~Lin, Z.~Long, Y.~Shen, M.~Zhao, Y.~Zhang, S.~Dong, X.~Wang, D.~Yin, L.~Ma \emph{et~al.}, ``Vita: Towards open-source interactive omni multimodal llm,'' \emph{arXiv preprint arXiv:2408.05211}, 2024.

\bibitem{freeze-omni}
X.~Wang, Y.~Li, C.~Fu, Y.~Shen, L.~Xie, K.~Li, X.~Sun, and L.~Ma, ``Freeze-omni: A smart and low latency speech-to-speech dialogue model with frozen llm,'' \emph{arXiv preprint arXiv:2411.00774}, 2024.

\bibitem{minmo}
Q.~Chen, Y.~Chen, Y.~Chen, M.~Chen, Y.~Chen, C.~Deng, Z.~Du, R.~Gao, C.~Gao, Z.~Gao \emph{et~al.}, ``Minmo: A multimodal large language model for seamless voice interaction,'' \emph{arXiv preprint arXiv:2501.06282}, 2025.

\bibitem{wavchat}
S.~Ji, Y.~Chen, M.~Fang, J.~Zuo, J.~Lu, H.~Wang, Z.~Jiang, L.~Zhou, S.~Liu, X.~Cheng \emph{et~al.}, ``Wavchat: A survey of spoken dialogue models,'' \emph{arXiv preprint arXiv:2411.13577}, 2024.

\bibitem{twist}
M.~Hassid, T.~Remez, T.~A. Nguyen, I.~Gat, A.~Conneau, F.~Kreuk, J.~Copet, A.~Defossez, G.~Synnaeve, E.~Dupoux \emph{et~al.}, ``Textually pretrained speech language models,'' \emph{Advances in Neural Information Processing Systems}, vol.~36, 2024.

\bibitem{spectron}
\BIBentryALTinterwordspacing
E.~Nachmani, A.~Levkovitch, R.~Hirsch, J.~Salazar, C.~Asawaroengchai, S.~Mariooryad, E.~Rivlin, R.~Skerry-Ryan, and M.~T. Ramanovich, ``Spoken question answering and speech continuation using spectrogram-powered {LLM},'' in \emph{The Twelfth International Conference on Learning Representations}, 2024. [Online]. Available: \url{https://openreview.net/forum?id=izrOLJov5y}
\BIBentrySTDinterwordspacing

\bibitem{gsqa}
M.-H. Shih, H.-L. Chung, Y.-C. Pai, M.-H. Hsu, G.-T. Lin, S.-W. Li, and H.~yi~Lee, ``Gsqa: An end-to-end model for generative spoken question answering,'' in \emph{Interspeech 2024}, 2024, pp. 2970--2974.

\bibitem{alignslm}
\BIBentryALTinterwordspacing
G.-T. Lin, P.~G. Shivakumar, A.~Gourav, Y.~Gu, A.~Gandhe, H.~yi~Lee, and I.~Bulyko, ``Align-slm: Textless spoken language models with reinforcement learning from ai feedback,'' 2024. [Online]. Available: \url{https://arxiv.org/abs/2411.01834}
\BIBentrySTDinterwordspacing

\bibitem{dual}
G.-T. Lin, Y.-S. Chuang, H.-L. Chung, S.~wen Yang, H.-J. Chen, S.~A. Dong, S.-W. Li, A.~Mohamed, H.~yi~Lee, and L.~shan Lee, ``{DUAL: Discrete Spoken Unit Adaptive Learning for Textless Spoken Question Answering},'' in \emph{Proc. Interspeech 2022}, 2022, pp. 5165--5169.

\bibitem{dynamic-superb}
C.-y. Huang, K.-H. Lu, S.-H. Wang, C.-Y. Hsiao, C.-Y. Kuan, H.~Wu, S.~Arora, K.-W. Chang, J.~Shi, Y.~Peng \emph{et~al.}, ``Dynamic-superb: Towards a dynamic, collaborative, and comprehensive instruction-tuning benchmark for speech,'' in \emph{ICASSP 2024-2024 IEEE International Conference on Acoustics, Speech and Signal Processing (ICASSP)}.\hskip 1em plus 0.5em minus 0.4em\relax IEEE, 2024, pp. 12\,136--12\,140.

\bibitem{dynamic-superb-2}
C.-y. Huang, W.-C. Chen, S.-w. Yang, A.~T. Liu, C.-A. Li, Y.-X. Lin, W.-C. Tseng, A.~Diwan, Y.-J. Shih, J.~Shi \emph{et~al.}, ``Dynamic-superb phase-2: A collaboratively expanding benchmark for measuring the capabilities of spoken language models with 180 tasks,'' \emph{arXiv preprint arXiv:2411.05361}, 2024.

\bibitem{voxdialogue}
\BIBentryALTinterwordspacing
X.~Cheng, R.~Hu, X.~Yang, J.~Lu, D.~Fu, Z.~Wang, S.~Ji, R.~Huang, B.~Zhang, T.~Jin, and Z.~Zhao, ``Voxdialogue: Can spoken dialogue systems understand information beyond words?'' in \emph{The Thirteenth International Conference on Learning Representations}, 2025. [Online]. Available: \url{https://openreview.net/forum?id=vbmSSIhKAM}
\BIBentrySTDinterwordspacing

\bibitem{sd-eval}
\BIBentryALTinterwordspacing
J.~Ao, Y.~Wang, X.~Tian, D.~Chen, J.~Zhang, L.~Lu, Y.~Wang, H.~Li, and Z.~Wu, ``{SD}-eval: A benchmark dataset for spoken dialogue understanding beyond words,'' in \emph{The Thirty-eight Conference on Neural Information Processing Systems Datasets and Benchmarks Track}, 2024. [Online]. Available: \url{https://openreview.net/forum?id=PnjbvbblGv}
\BIBentrySTDinterwordspacing

\bibitem{air-bench}
Q.~Yang, J.~Xu, W.~Liu, Y.~Chu, Z.~Jiang, X.~Zhou, Y.~Leng, Y.~Lv, Z.~Zhao, C.~Zhou \emph{et~al.}, ``Air-bench: Benchmarking large audio-language models via generative comprehension,'' \emph{arXiv preprint arXiv:2402.07729}, 2024.

\bibitem{styletalk}
\BIBentryALTinterwordspacing
G.-T. Lin, C.-H. Chiang, and H.-y. Lee, ``Advancing large language models to capture varied speaking styles and respond properly in spoken conversations,'' in \emph{Proceedings of the 62nd Annual Meeting of the Association for Computational Linguistics (Volume 1: Long Papers)}, L.-W. Ku, A.~Martins, and V.~Srikumar, Eds.\hskip 1em plus 0.5em minus 0.4em\relax Bangkok, Thailand: Association for Computational Linguistics, Aug. 2024, pp. 6626--6642. [Online]. Available: \url{https://aclanthology.org/2024.acl-long.358}
\BIBentrySTDinterwordspacing

\bibitem{paralingpt}
G.-T. Lin, P.~G. Shivakumar, A.~Gandhe, C.-H.~H. Yang, Y.~Gu, S.~Ghosh, A.~Stolcke, H.-Y. Lee, and I.~Bulyko, ``Paralinguistics-enhanced large language modeling of spoken dialogue,'' in \emph{ICASSP 2024 - 2024 IEEE International Conference on Acoustics, Speech and Signal Processing (ICASSP)}, 2024, pp. 10\,316--10\,320.

\bibitem{emphasis}
\BIBentryALTinterwordspacing
G.-T. Lin and H.-y. Lee, ``Can {LLM}s understand the implication of emphasized sentences in dialogue?'' in \emph{Findings of the Association for Computational Linguistics: EMNLP 2024}, Y.~Al-Onaizan, M.~Bansal, and Y.-N. Chen, Eds.\hskip 1em plus 0.5em minus 0.4em\relax Miami, Florida, USA: Association for Computational Linguistics, Nov. 2024, pp. 13\,391--13\,401. [Online]. Available: \url{https://aclanthology.org/2024.findings-emnlp.782/}
\BIBentrySTDinterwordspacing

\bibitem{dglsm}
T.~A. Nguyen, E.~Kharitonov, J.~Copet, Y.~Adi, W.-N. Hsu, A.~Elkahky, P.~Tomasello, R.~Algayres, B.~Sagot, A.~Mohamed \emph{et~al.}, ``Generative spoken dialogue language modeling,'' \emph{Transactions of the Association for Computational Linguistics}, vol.~11, pp. 250--266, 2023.

\bibitem{fsm}
\BIBentryALTinterwordspacing
P.~Wang, S.~Lu, Y.~Tang, S.~Yan, W.~Xia, and Y.~Xiong, ``A full-duplex speech dialogue scheme based on large language model,'' in \emph{The Thirty-eighth Annual Conference on Neural Information Processing Systems}, 2024. [Online]. Available: \url{https://openreview.net/forum?id=YawXY6mWiK}
\BIBentrySTDinterwordspacing

\bibitem{minicpm-duplex}
\BIBentryALTinterwordspacing
X.~Zhang, Y.~Chen, S.~Hu, X.~Han, Z.~Xu, Y.~Xu, W.~Zhao, M.~Sun, and Z.~Liu, ``Beyond the turn-based game: Enabling real-time conversations with duplex models,'' in \emph{Proceedings of the 2024 Conference on Empirical Methods in Natural Language Processing}, Y.~Al-Onaizan, M.~Bansal, and Y.-N. Chen, Eds.\hskip 1em plus 0.5em minus 0.4em\relax Miami, Florida, USA: Association for Computational Linguistics, Nov. 2024, pp. 11\,543--11\,557. [Online]. Available: \url{https://aclanthology.org/2024.emnlp-main.644/}
\BIBentrySTDinterwordspacing

\bibitem{parrot}
\BIBentryALTinterwordspacing
Q.~Wang, Z.~Meng, W.~Cui, Y.~Zhang, P.~Wu, B.~Wu, Z.~Zheng, I.~King, L.~Chen, and P.~Zhao, ``Parrot: Seamless spoken dialogue interaction with double-channel large language models,'' 2025. [Online]. Available: \url{https://openreview.net/forum?id=73EDGbG6mB}
\BIBentrySTDinterwordspacing

\bibitem{minicpm-duo}
W.~Xu, S.~Wang, W.~Zhao, X.~Han, Y.~Yan, Y.~Zhang, Z.~Tao, Z.~Liu, and W.~Che, ``Enabling real-time conversations with minimal training costs,'' \emph{arXiv preprint arXiv:2409.11727}, 2024.

\bibitem{salmonn-omni}
W.~Yu, S.~Wang, X.~Yang, X.~Chen, X.~Tian, J.~Zhang, G.~Sun, L.~Lu, Y.~Wang, and C.~Zhang, ``Salmonn-omni: A codec-free llm for full-duplex speech understanding and generation,'' \emph{arXiv preprint arXiv:2411.18138}, 2024.

\bibitem{omniflatten}
Q.~Zhang, L.~Cheng, C.~Deng, Q.~Chen, W.~Wang, S.~Zheng, J.~Liu, H.~Yu, C.~Tan, Z.~Du \emph{et~al.}, ``Omniflatten: An end-to-end gpt model for seamless voice conversation,'' \emph{arXiv preprint arXiv:2410.17799}, 2024.

\bibitem{rttldg}
L.~Mai and J.~Carson-Berndsen, ``Real-time textless dialogue generation,'' \emph{arXiv preprint arXiv:2501.04877}, 2025.

\bibitem{nova-sonic}
A.~A.~G. Intelligence, ``Amazon nova sonic: Technical report and model card,'' 2025.

\bibitem{talking-turns}
\BIBentryALTinterwordspacing
S.~Arora, Z.~Lu, C.-C. Chiu, R.~Pang, and S.~Watanabe, ``Talking turns: Benchmarking audio foundation models on turn-taking dynamics,'' in \emph{The Thirteenth International Conference on Learning Representations}, 2025. [Online]. Available: \url{https://openreview.net/forum?id=2e4ECh0ikn}
\BIBentrySTDinterwordspacing

\bibitem{superb}
S.~wen Yang, P.-H. Chi, Y.-S. Chuang, C.-I.~J. Lai, K.~Lakhotia, Y.~Y. Lin, A.~T. Liu, J.~Shi, X.~Chang, G.-T. Lin, T.-H. Huang, W.-C. Tseng, K.~tik Lee, D.-R. Liu, Z.~Huang, S.~Dong, S.-W. Li, S.~Watanabe, A.~Mohamed, and H.~yi~Lee, ``{SUPERB: Speech Processing Universal PERformance Benchmark},'' in \emph{Proc. Interspeech 2021}, 2021, pp. 1194--1198.

\bibitem{superb-prosody}
G.-T. Lin, C.-L. Feng, W.-P. Huang, Y.~Tseng, T.-H. Lin, C.-A. Li, H.-y. Lee, and N.~G. Ward, ``On the utility of self-supervised models for prosody-related tasks,'' in \emph{Proc.\ IEEE SLT}, 2023, pp. 1104--1111.

\bibitem{superb-sg}
H.-S. Tsai, H.-J. Chang, W.-C. Huang, Z.~Huang, K.~Lakhotia, S.-w. Yang, S.~Dong, A.~Liu, C.-I. Lai, J.~Shi \emph{et~al.}, ``Superb-sg: Enhanced speech processing universal performance benchmark for semantic and generative capabilities,'' in \emph{Proceedings of the 60th Annual Meeting of the Association for Computational Linguistics (Volume 1: Long Papers)}, 2022, pp. 8479--8492.

\bibitem{voicebench}
Y.~Chen, X.~Yue, C.~Zhang, X.~Gao, R.~T. Tan, and H.~Li, ``Voicebench: Benchmarking llm-based voice assistants,'' \emph{arXiv preprint arXiv:2410.17196}, 2024.

\bibitem{zerospeech}
T.~A. Nguyen, M.~de~Seyssel, P.~Roz{\'e}, M.~Rivi{\`e}re, E.~Kharitonov, A.~Baevski, E.~Dunbar, and E.~Dupoux, ``The zero resource speech benchmark 2021: Metrics and baselines for unsupervised spoken language modeling,'' in \emph{NeuRIPS Workshop on Self-Supervised Learning for Speech and Audio Processing}, 2020.

\bibitem{storycloze}
\BIBentryALTinterwordspacing
N.~Mostafazadeh, M.~Roth, A.~Louis, N.~Chambers, and J.~Allen, ``{LSDS}em 2017 shared task: The story cloze test,'' in \emph{Proceedings of the 2nd Workshop on Linking Models of Lexical, Sentential and Discourse-level Semantics}, M.~Roth, N.~Mostafazadeh, N.~Chambers, and A.~Louis, Eds.\hskip 1em plus 0.5em minus 0.4em\relax Valencia, Spain: Association for Computational Linguistics, Apr. 2017, pp. 46--51. [Online]. Available: \url{https://aclanthology.org/W17-0906}
\BIBentrySTDinterwordspacing

\bibitem{ward}
N.~Ward and W.~Tsukahara, ``Prosodic features which cue back-channel responses in english and japanese,'' \emph{Journal of pragmatics}, vol.~32, no.~8, pp. 1177--1207, 2000.

\bibitem{reece2023candor}
A.~Reece, G.~Cooney, P.~Bull, C.~Chung, B.~Dawson, C.~Fitzpatrick, T.~Glazer, D.~Knox, A.~Liebscher, and S.~Marin, ``The candor corpus: Insights from a large multimodal dataset of naturalistic conversation,'' \emph{Science Advances}, vol.~9, no.~13, p. eadf3197, 2023.

\bibitem{jefferson1989preliminary}
G.~Jefferson, ``Preliminary notes on a possible metric which provides for a ‘standard maximum’silence of approximately one second in conversation,'' \emph{Conversation: An Interdisciplinary Approach/Multilingual Matters}, 1989.

\bibitem{liu2022pause}
S.~Liu, Y.~Nakajima, L.~Chen, S.~Arndt, M.~Kakizoe, M.~A. Elliott, and G.~B. Remijn, ``How pause duration influences impressions of english speech: Comparison between native and non-native speakers,'' \emph{Frontiers in psychology}, vol.~13, p. 778018, 2022.

\bibitem{heldner2010pauses}
M.~Heldner and J.~Edlund, ``Pauses, gaps and overlaps in conversations,'' \emph{Journal of Phonetics}, vol.~38, no.~4, pp. 555--568, 2010.

\bibitem{icc}
\BIBentryALTinterwordspacing
M.~Umair, V.~Sarathy, and J.~Ruiter, ``Large language models know what to say but not when to speak,'' in \emph{Findings of the Association for Computational Linguistics: EMNLP 2024}, Y.~Al-Onaizan, M.~Bansal, and Y.-N. Chen, Eds.\hskip 1em plus 0.5em minus 0.4em\relax Miami, Florida, USA: Association for Computational Linguistics, Nov. 2024, pp. 15\,503--15\,514. [Online]. Available: \url{https://aclanthology.org/2024.findings-emnlp.909/}
\BIBentrySTDinterwordspacing

\bibitem{gpt4}
OpenAI, ``Gpt-4 technical report,'' 2023.

\bibitem{hubert}
W.-N. Hsu, B.~Bolte, Y.-H.~H. Tsai, K.~Lakhotia, R.~Salakhutdinov, and A.~Mohamed, ``Hubert: Self-supervised speech representation learning by masked prediction of hidden units,'' \emph{IEEE/ACM transactions on audio, speech, and language processing}, vol.~29, pp. 3451--3460, 2021.

\bibitem{cieri2004fisher}
C.~Cieri, D.~Graff, O.~Kimball, D.~Miller, and K.~Walker, ``Fisher english training speech part 1 transcripts,'' \emph{Philadelphia: Linguistic Data Consortium}, 2004.

\end{thebibliography}

% \newpage
% \mbox{}  % 空白页
% \newpage
% \input{appendix}

\end{document}